\documentclass[twocolumn]{article}
\usepackage[utf8]{inputenc}
\usepackage[T1]{fontenc}    % use 8-bit T1 fonts
\usepackage{authblk}
\usepackage{setspace}
\usepackage{graphicx}
\usepackage[switch]{lineno}
\usepackage{xcolor}
\usepackage{listings}  % For typesetting code
\usepackage{inconsolata}  % Nicer \texttt font
\usepackage{booktabs}
\usepackage{longtable}
\usepackage[textsize=footnotesize]{todonotes}
\usepackage{verbatim}
\usepackage{xspace}
\usepackage{abstract}
\usepackage{multirow}
\usepackage{dirtree}

\usepackage[colorlinks=true, linkcolor=blue, citecolor=blue, urlcolor=blue]{hyperref}
\usepackage[capitalize]{cleveref}

\usepackage{sectsty}
\subsectionfont{\sffamily\normalsize\raggedright\MakeUppercase}

\usepackage{caption}
\crefname{listing}{Code Snippet}{Code Snippets}

\newcommand{\datasetname}{PadChest-GR\xspace}
\newcommand{\dataseturl}{\url{https://bimcv.cipf.es/bimcv-projects/padchest-gr/}}

\newcommand{\groundedreportgeneration}{GRRG\xspace}

\title{\vspace*{-2cm}\datasetname: A Bilingual Chest X-ray Dataset for\\Grounded Radiology Report Generation}

\author[1]{Daniel Coelho de Castro}  % PhD, dacoelh@microsoft.com
\author[2]{Aurelia Bustos} % PhD, aurelia@medbravo.org
\author[1]{Shruthi Bannur}  % MSc, shruthi.bannur@microsoft.com
\author[1]{Stephanie L. Hyland}  % PhD, stephanie.hyland@microsoft.com
\author[1]{\authorcr Kenza Bouzid}  % MSc, kenza.bouzid@microsoft.com
\author[3,1]{Maria Teodora Wetscherek}  % MD PhD, teo.wetscherek@nhs.net
\author[4]{Maria Dolores Sánchez-Valverde} % MD, masavalverde@gmail.com
\author[4]{\authorcr Lara Jaques-Pérez} % MD, jaques_lar@gva.es
\author[4]{Lourdes Pérez-Rodríguez} % MD, perez_lourdesrod@gva.es
\author[1]{Kenji Takeda}  % PhD, kenji.takeda@microsoft.com
\author[5]{José María Salinas}  % PhD, salinas_josser@gva.es
\author[1]{\authorcr Javier Alvarez-Valle}  % MSc, jaalvare@microsoft.com
\author[4]{Joaquín Galant-Herrero} % PhD, galant_joa@gva.es
\author[6]{Antonio Pertusa} % PhD, pertusa@ua.es

\affil[1]{Microsoft Research, Cambridge, UK}
\affil[2]{Medbravo, Alicante, Spain}
\affil[3]{Department of Radiology, University of Cambridge and Cambridge University Hospitals NHS Foundation Trust, Cambridge, UK}
\affil[4]{Department of Radiology, University Hospital Sant Joan d'Alacant, Spain}
\affil[5]{Department of Health Informatics, University Hospital Sant Joan d'Alacant, Spain}
\affil[6]{University Institute for Computing Research, University of Alicante, Spain}

\date{}  % Remove date

\newcommand{\abssec}[1]{{\sffamily\textbf{#1}}}

\begin{document}

% %TC:ignore
% % \section*{[Word counts]}
% Submission guide: \url{https://ai.nejm.org/author-center/article-types-and-submission-information}
% \newline
% Abstract (Backgrounds, Methods, Results, Conclusion): 300 words maximum\newline
% 1-2 sentence description\newline
% Maximum words: \textbf{3,000}\newline
% Up to \textbf{5 tables and figures}\newline
% Up to \textbf{40 references}\newline
% Authors: no limit\newline

% \detailtexcount{main}
% \clearpage

%\linenumbers
\onehalfspacing

\twocolumn[
\maketitle
%TC:endignore

% Abstract: Max 300 words
% Count words in abstract: https://tex.stackexchange.com/a/354070
%TC:break Abstract

\vspace{-1cm}

\begin{onecolabstract}

\abssec{BACKGROUND} 
% v1
Radiology report generation (RRG) aims to create free-text radiology reports from clinical imaging. Grounded radiology report generation (\groundedreportgeneration) extends RRG by including the localisation of individual findings on the image. Currently, there are no manually annotated  chest X-ray (CXR) datasets to train \groundedreportgeneration models.

\abssec{METHODS}
In this work, we present a dataset called \datasetname (Grounded-Reporting) derived from PadChest aimed at training \groundedreportgeneration models for CXR images. First, a subset of studies was selected from PadChest using images with frontal projection, excluding paediatric patients and studies originally labelled as sub-optimal. 
Then, using GPT-4 in Microsoft Azure OpenAI Service, reports were processed to extract single-finding sentences, translate them from Spanish into English, link them to the existing PadChest finding and location labels, and classify finding progression.
A team of 14 radiologists reviewed and manually annotated the findings in each image using bounding boxes, first discarding some studies with issues on the image quality, report, or findings list, and then annotating the boxes for each finding.
% Since annotation can be subjective, we provide the bounding boxes proposed by two radiologists for each image, prioritising the choice made by the most senior radiologist. 

\abssec{RESULTS}
We curate a public bi-lingual dataset of 4,555 CXR studies with grounded reports (3,099 abnormal and 1,456 normal), each containing complete lists of sentences describing individual present (positive)  and absent (negative) findings in English and Spanish. In total, \datasetname contains 7,037 positive and 3,422 negative finding sentences. Every positive finding sentence is associated with up to two independent sets of bounding boxes labelled by different readers and has categorical labels for finding type, locations, and progression.

%\wip{To do: Add number of images, findings, boxes. Training/test splits, stats}

\abssec{CONCLUSIONS} To the best of our knowledge, \datasetname is the first manually curated dataset designed to train \groundedreportgeneration models for understanding and interpreting radiological images and generated text. By including detailed localization and comprehensive annotations of all clinically relevant findings, it provides a valuable resource for developing and evaluating \groundedreportgeneration models from CXR images. \datasetname can be downloaded under request from \dataseturl.

%As discussed, \datasetname also faces limitations such as regional bias, lack of lateral views, and inter-observer variability, but future efforts could address these issues with the experience learned by our multi-disciplinary team. \datasetname can be downloaded from \dataseturl.

\end{onecolabstract}
%TC:break main
]

\pagenumbering{arabic}

\clearpage

\section*{Introduction}

% Proposed outline (~4 paragraphs):
% - Broader context: (chest) radiology, use of AI/ML for draft reporting
% - Problem setting: Importance of spatial grounding + definition of grounded reporting
% - Literature gap: Existing datasets with localisation + limitations w.r.t. grounded reporting
% - Aims/Contributions: Overview of PadChest-GR + novelty (grounded reports, use of LLMs)

%Among the various imaging modalities for diagnosis and treatment, X-ray imaging is widely used due to its low acquisition time, cost-effectiveness, and ability to provide valuable findings. 

\begin{figure*}[tp]
    \centering
    % \missingfigure{Top panel: original sample from PadChest with current and prior image, Spanish report, and finding+location labels}
    \includegraphics[width=\linewidth]{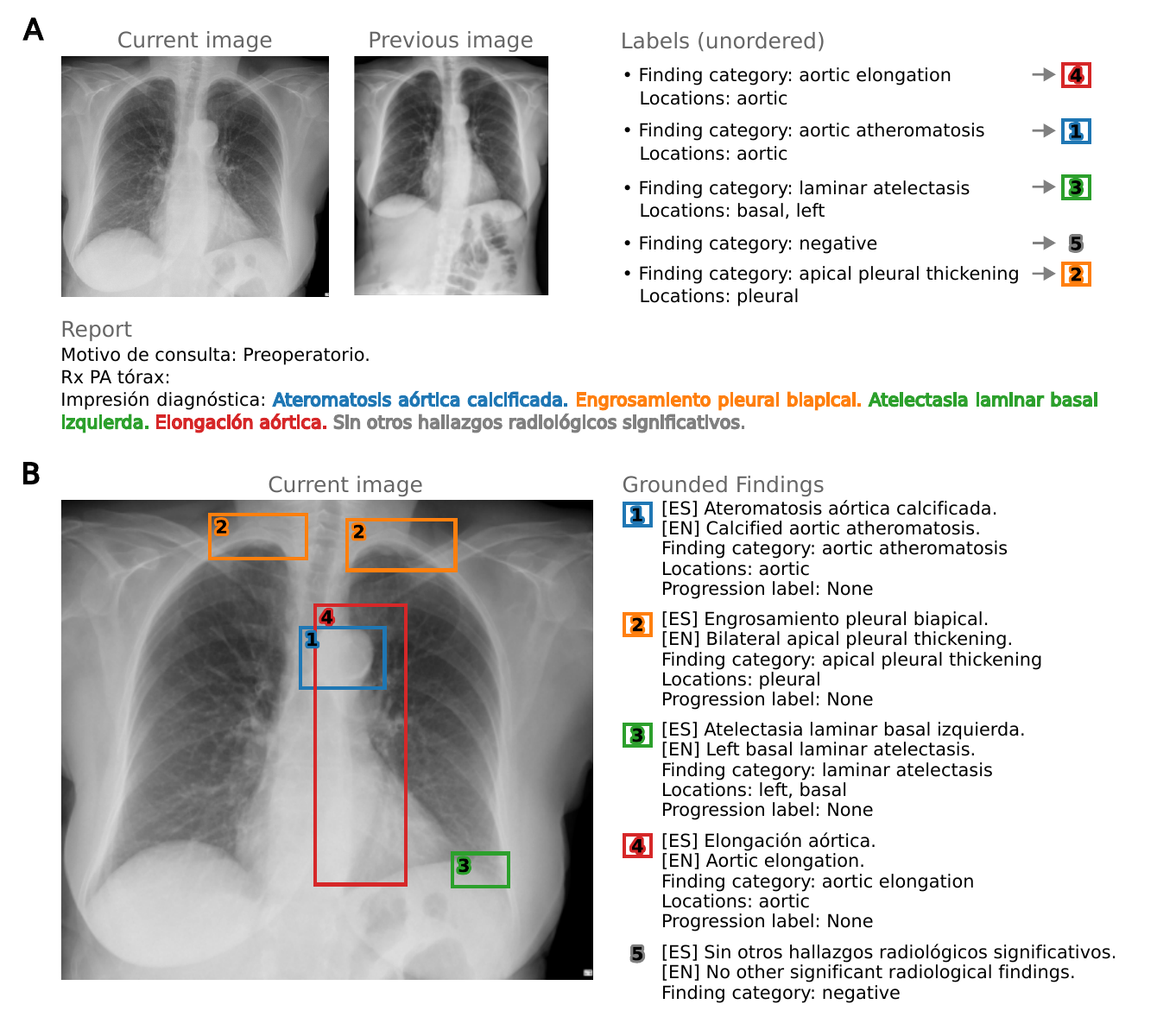}
    \caption{Example of a grounded report from \datasetname.
    (Panel~A)~An original sample from PadChest \cite{bustos2019padchest}, including a chest X-ray image, its free-text radiology report in Spanish, and categorical labels for radiological findings and anatomical locations (originally in English). We also show an image from the patient's most recent prior study for comparison.
    (Panel~B)~The corresponding sample in \datasetname. The image is accompanied with a report in the form of a list of sentences individually describing at most one finding, in both English and Spanish. For findings that pertain to specific regions of the image, bounding boxes are provided to indicate the location. A finding can have more than one bounding box, for example in the case of bilateral observations.
    This comprehensive set of annotations for all finding sentences differs from existing datasets, which associate bounding boxes with only categorical labels or single sentences.
    }
    \label{fig:main_example}
    % Figure metadata
    %Study: 244704331965478346516087003646856493817
    % Current image: images/244704331965478346516087003646856493817_w4wxrp.png
    % Prior image: images/216840111366964012373310883942009170112608544_00-096-127.png
\end{figure*}

In recent years, the use of artificial intelligence (AI) to improve medical image analysis has gained significant interest, with the potential to alleviate radiology workloads and enhance patient care \cite{rajpurkar2023current, yildirim2024multimodal}.
% creating a demand for high-quality annotated datasets.
%Radiologists receiving AI assistance can achieve better performance than unassisted clinicians \cite{ahn2022association}, but as pointed out by \cite{rajpurkar2023current}, human–AI collaboration for image interpretation offers mixed evidence regarding the value of such a collaboration.
The modelling task of \groundedreportgeneration can be defined as predicting a list of sentences or phrases describing all individual findings in an image, with associated spatial annotations (e.g.~bounding boxes) for localisable findings \cite{bannur2024maira2}.
An example grounded report is shown in \cref{fig:main_example}.
% As seen in \cref{fig:main_example}, a grounded report\cite{bannur2024maira2} is a description of all findings in an image with accompanying localisation.

Because explainability is crucial in this domain \cite{rajpurkar2023current}, spatially grounding radiological findings is expected to help with verification of AI-generated draft reports \cite{bernstein2023incorrect}.
Such grounding can also underpin new interactive capabilities of medical AI models \cite{moor2023foundation}, facilitating their interpretation by clinicians and even patients \cite{yildirim2024multimodal}.

In the literature, there exist many CXR image datasets labelled for diagnosis and finding classification tasks \cite{bustos2019padchest, irvin2019chexpert, radgraphxl, wang2017chestxray8}, or accompanied by textual radiology reports for automated draft report generation \cite{johnson2019mimiccxr, demner2016iuxray, feng2021candidptx, chambon2024chexpertplus}.
Some datasets also include spatial annotations to localise labels (for finding, anatomy, or device; e.g.~\textit{`pneumothorax'}) \cite{Wu20,Wu21, Lanfredi22,vindr, wang2017chestxray8, feng2021candidptx} or single finding phrases \cite{boecking2022mscxr,bannur2023mscxrt}, such as \textit{`Left retrocardiac opacity'}.
% To the best of our knowledge, there are no datasets manually annotated for grounded report generation. 
% While such datasets enable the development of models for object detection and phrase grounding, they are not sufficient to support the generation of full grounded reports.
However, such datasets are not sufficient to construct full grounded reports, as they lack spatial annotations linked to complete sets of descriptive finding sentences
\cite{ocae202}.
%{sec:results} shows the results and statistics of the curated dataset. Fin%r. Download at \url{https://b2drop.bsc.es/index.php/s/PadChest-GR}
%\todo[inline]{Add summary paragraph and link to Figure XXX.(Antonio: Summary not necessary in this journal, ref to figure at the header of next section (it's allowed))}

\section*{Methods}
\label{sec:methods}

\begin{figure*}[t]
    \centering
    % trim={<left> <lower> <right> <upper>}
    \includegraphics[trim={5cm 5cm 27cm 5cm},clip,width=\textwidth]{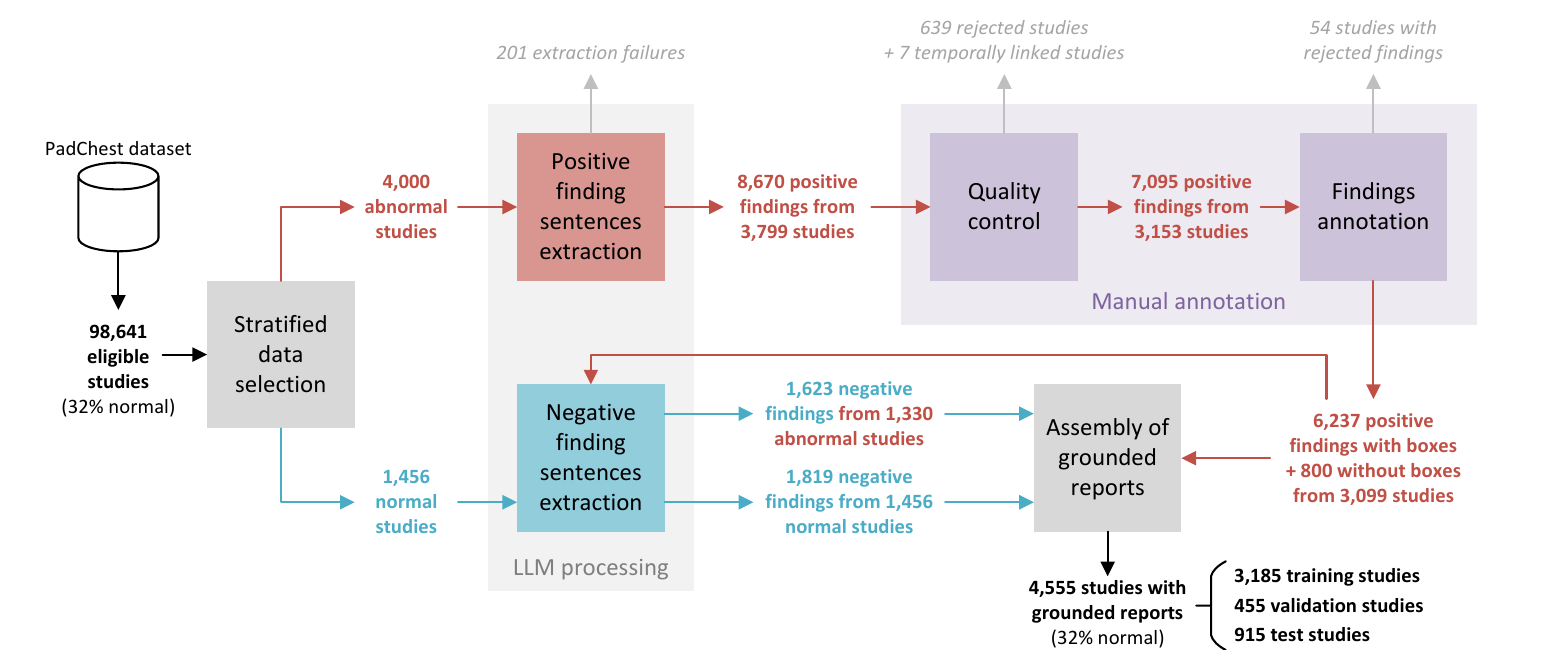}
    \caption{Illustration of the data curation pipeline.
    From the eligible PadChest studies, a set of studies with positive (i.e.~abnormal) findings were selected with stratified sampling using a prespecified list of common and relevant finding categories.
    Normal studies (those with no positive findings) were also sampled to maintain the proportion of normal studies in the overall PadChest dataset (32.0\%).
    From the findings section of each study, large language model (LLM) processing extracted and classified individual finding sentences (both positive and negative), and translated them to English.
    Positive findings were manually annotated by a team of radiologists to conduct both quality control and annotation of bounding boxes.
    Positive and negative findings sentences were assembled with these annotations in the original order for each study to produce a fully grounded report.
    }
    \label{fig:data-flow}
\end{figure*}

\subsection*{Source data}
PadChest \cite{bustos2019padchest} is a large-scale CXR dataset which includes more than 160,000 images obtained from 67,000 patients that were interpreted and reported by radiologists at the University Hospital Sant Joan d'Alacant (Spain) from 2009 to 2017.  The reports were labeled with 174 different radiological findings, 19 differential diagnoses, and 104 anatomical locations organized as a hierarchical taxonomy and mapped onto standard Unified Medical Language System (UMLS) terminology \cite{bodenreider2004umls}.
The text reports publicly released with PadChest had been heavily processed to preserve anonymization, normalise capitalisation, punctuation, accents, stop-words, and affixes. For \datasetname, we retrieved the original reports for the entire PadChest cohort after anonymization and without further processing. %  were already preprocessed to preserve anonymization, and which we release along with \datasetname.

% Copying some info from the selection criteria document for reference

%%%% Technical exclusion criteria
% - Try to link to prior image based on DICOM timestamp
% - Drop any studies with unloadable images (whether the unloadable images were in the current or prior study)
% - We observe studies and patients with different IDs having the same report. We drop all such duplicate studies / patients in the dataset (also dropping any studies that have these as priors)
% - Drop any studies with time difference == 0 to the prior, since we can’t tell the ordering of current and prior study. Also drop any studies with prior study pointing to one of these studies
% - Drop images marked ``suboptimal'' (whether the current or prior study)
% - Drop any paediatric studies (we drop the full patient)
% - Filter to just AP / AP_horizontal / PA images. Drop any studies where this filtering causes the prior study to no longer have any images.
% - Filter out any studies with > 1 frontal image. Also drop any studies with >1 prior frontal image
% - Drop images marked ``exclude'' 
% - Drop any report with the label ``unchanged''
% - Filter any studies with empty reports
% - Filter out normal studies
% - Filter to manual labels
For inclusion in \datasetname, each radiographic study in PadChest was first linked to the patient's most recent prior study based on the \texttt{StudyDate\_DICOM} field. Studies from the same patient on the same date were excluded, as their order cannot be determined.
We selected only frontal images (erect or supine, with projection  \textit{`PA'}, \textit{`AP'}, or \textit{`AP-horizontal'}) and excluded the few studies with more than one frontal image in either the current or prior study to avoid ambiguity. 
Paediatric patients (18 years old or younger) were also rejected, as well as studies acquired with a paediatric protocol. 
We also filtered out any studies that included the \textit{`suboptimal'}, \textit{`exclude'}, or \textit{`unchanged'} labels in PadChest, as they cannot be reliably annotated.
% \wip{Lastly, note that approximately 25\% of PadChest samples were originally labelled by radiologists, then the remaining cases were annotated automatically by a model trained on the manual labels \cite{bustos2019padchest}. For \datasetname, we included only the more reliable manually labelled subset.}\todo{Decide whether to mention this here as part of the eligibility criteria or later within the selection process}
\Cref{fig:data-flow} summarises the pipeline for constructing \datasetname out of the 98,641 eligible PadChest studies.
% \todo{Update counts to reflect only manually labelled studies}

\subsection*{Stratified data selection}
We selected 4,000 abnormal studies (i.e.~with at least one positive finding), chosen randomly via stratified sampling according to the distribution of finding categories in the eligible cohort.
Note that approximately 25\% of PadChest samples were originally labelled by radiologists, whereas the remaining cases were annotated automatically by a recurrent neural network with a per-label attention mechanism \cite{mullenbach2018explainable}. For \datasetname, we sampled only from the more reliable manually labelled subset.

Further, we highlight that the original PadChest finding ontology is too granular for stratification and analysis, as many detailed findings are rare in the dataset and could not be faithfully represented in the much smaller \datasetname dataset. 
Therefore, we first enumerated a non-exhaustive subset of finding types that were clinically relevant, broad enough for significant prevalence but not too vague, and visible on a CXR. 
Finer-grained findings were subsumed into their parent super-categories (if selected) following the original PadChest hierarchy \cite{bustos2019padchest}; see full mapping in \cref{tab:group_labels}.
The final list of findings was further validated by a panel of radiologists.
% All categories that remained unmapped or had lower than 1\% prevalence after this grouping were mapped to \textit{`Other'} for analysis purposes (see \cref{tab:other_entities}).
All categories that remained unmapped or had lower than 1\% prevalence after this grouping were mapped to \textit{`Other'} (see \cref{tab:other_entities}) and were simply not tracked for stratification, but the corresponding samples could still be included at random.

%%%% Selection of normal cases
Finally, to preserve the composition of the original cohort, the dataset was complemented with normal studies---i.e.~those with no positive findings according to the PadChest labels. After annotation and quality control of the abnormal cases (as described later in `Manual annotation' section), we randomly selected enough eligible normal studies to match their overall prevalence of 32.0\%.

\subsection*{Extraction of finding sentences}
% In the task of grounded radiology reporting, each observed finding should be associated with spatial annotations (e.g.~one or more bounding boxes) describing their location in an image \cite{bannur2024maira2}.
Sentences in routine free-text radiology reports frequently refer to multiple findings, and may also mention external information (e.g.~patient history), communications between healthcare staff, clinical interpretation, follow-up recommendations, etc.---which cannot be objectively inferred from the radiographic study alone.
% Therefore, to enable grounding annotation and development of grounded reporting models, existing radiology datasets must be processed to accurately extract sentences describing individual objective findings.
On the other hand, grounded radiology reporting involves separately localising each observed finding \cite{bannur2024maira2}.
Therefore, to enable annotation and model development for this task, existing radiology datasets must be processed to accurately extract sentences describing individual objective findings.

%%%% Positive findings
 report in Spanish was processed to not only extract the single-finding sentences, but also translate the latter to English and link them to the existing PadChest labels.()
See details and model instruction prompts in \cref{app:llm_details}.
Positive and negative finding phrases were extracted in two separate stages.

% - Sentence-level extraction based on finding/location labels
% - Verbatim sentence copying to enable linkage
% - Progression classification
% - Translation to English
For the positive findings, we leveraged PadChest's existing finding and location labels, which were originally collected at the sentence level \cite{bustos2019padchest}. Labels for differential diagnoses that rely on additional clinical context, such as \textit{`pneumonia'}, were ignored.
The model was presented with the full original report and grouped sentence labels, and was asked to generate standalone sentences describing exclusively each of the given finding labels, along with English translations.
Every extracted finding sentence was also matched to the relevant subset of provided location labels (see \cref{sec:locations}). Further, the model classified the temporal progression of each finding, when applicable (see \cref{fig:progression}).
% The full prompt with detailed instructions is given in \cref{lst:positive_findings_prompt}.

%%%% Negative findings
Negative findings are statements referring to the absence of abnormal observations, such as \textit{`No evidence of consolidation'}, \textit{`Heart size is normal'}, or \textit{`No significant findings'}. 
In PadChest, negative findings were not labelled for the specific negated findings or their locations \cite{bustos2019padchest}.
% In PadChest, negative findings were labelled at study level as \textit{`normal'} if there were no positive findings reported, without reference to the negated findings or their locations \cite{bustos2019padchest}.
% Unlike for positive findings, in PadChest the negation of findings was summarized at the report level with the negative label `normal' and  hence, it did not include the reference to the negated findings or their localizations.
% Specifically, in PadChest, sentences describing normality, i.e.~those that either reported complete resolution, did not describe radiographic findings or negated their presence were labeled as `normal'.
% In a second step, the report was assigned a single label `normal' only if in addition to that label, there were no other labels regarded as a radiographic finding or diagnoses.
We therefore relied on the LLM to identify negative finding sentences and split them if necessary. Following \cite{bannur2024maira2}, we tasked the LLM with extracting sentences describing individual observations, with two additional requirements: (1)~translation to English, and (2)~classification of sentences as either positive or negative findings. This enables us to extract only the negative findings.
% Eleven in-context examples were provided.
% The task instruction and one in-context example are presented in \cref{lst:negative_findings_prompt,lst:negative_findings_fewshot}.

%%%% QC and clean-up
% \todo[inline]{Possibly describe any final QC and clean-up steps, and assembly of full grounded reports}

% \begin{figure*}[p]
% \centering
% \includegraphics[width=0.7\textwidth]{figures/task1sample.png}
% \includegraphics[width=0.7\textwidth]{figures/task4sample.png}
% \caption{Screenshots of the interface for both manual annotation stages.
% (Panel~A)~Example of study-level quality control, containing bilateral apical thickening, interstitial-alveolar infiltrate, and diffuse interstitial pattern. In this stage, the radiologists select the category (for this study, `no issue') according to the quality of the image, the report, and the extracted findings.
% (Panel~B)~Example of bounding box annotation for an individual finding (interstitial-alveolar infiltrate) from the same study.
% All annotations were performed using the Centaur Labs platform.
% }
% \label{fig:annotation_interface}
% \todo[inline]{@Antonio: Update screenshots to a sample with prior. Answered: There is no prior available for this study. We could select a more proper image with prior, but this should be done by a MD.}
% \todo[inline]{Make these into proper subfigure panels}
% \end{figure*}

\begin{figure*}[p]
\centering
\includegraphics[width=0.8\textwidth]{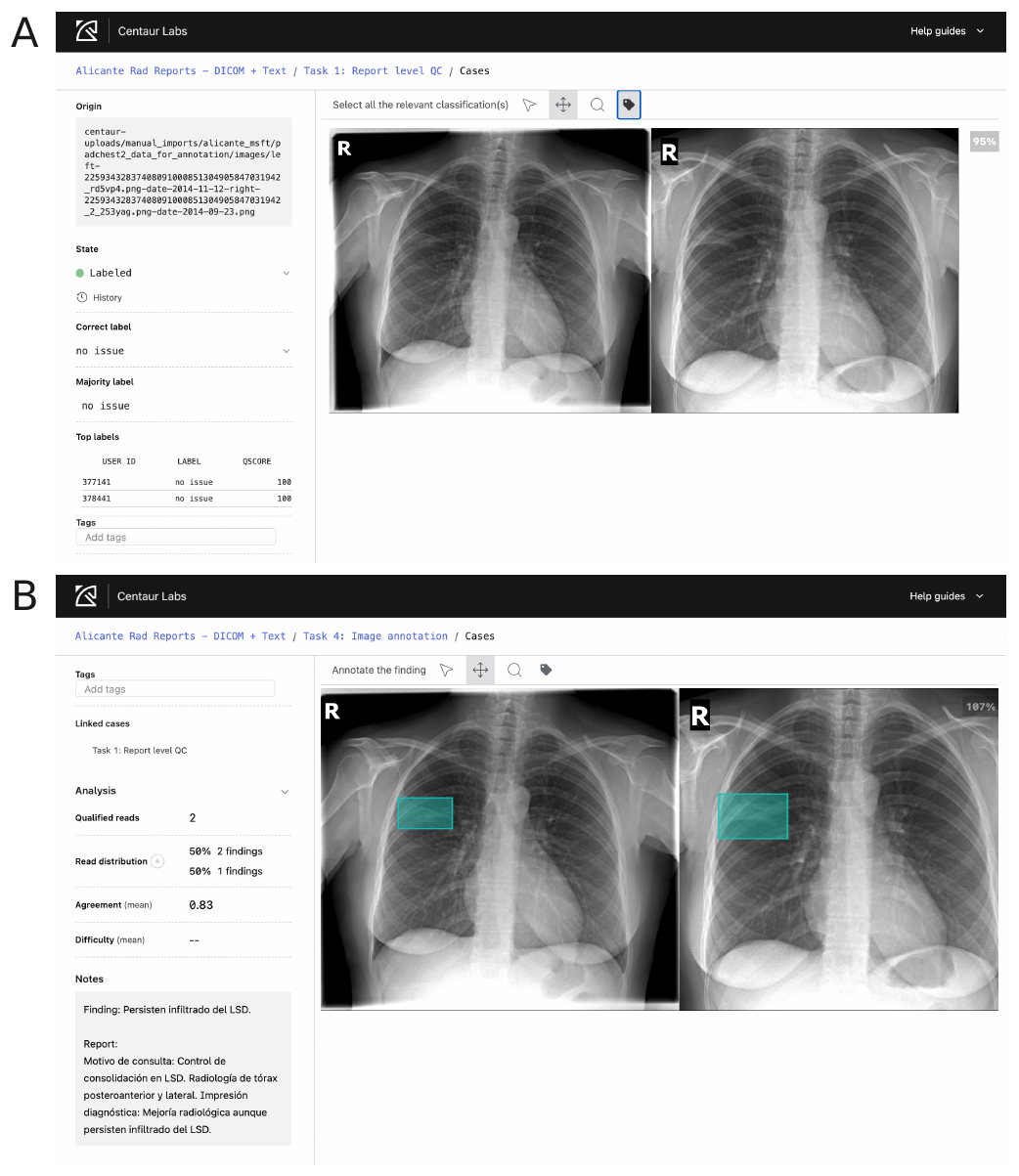}
\caption{Screenshots of the interface for both manual annotation stages.
(Panel~A)~Example of study-level quality control, containing infiltrates in the current (left) and prior (right) images. In this stage, the radiologists select the category (for this study, \textit{`no issue'}) according to the quality of the image, the report, and the extracted findings.
(Panel~B)~Example of bounding box annotation for an individual finding (\textit{`Infiltrates in the right upper lobe persist'}) from the same study.
All annotations were performed using the Centaur Labs platform.
}
\label{fig:annotation_interface}
\end{figure*}

\subsection*{Manual annotation}

The selected studies were annotated by nine senior and five junior radiologists from the University Hospital Sant Joan d'Alacant, Spain, using the HIPAA-compliant Centaur Labs labelling service  (\url{https://www.centaurlabs.com}).  
The annotation was performed in two stages: study-level quality control followed by box annotations for each extracted positive finding.
For both stages, every study or finding was analysed independently by two professionals.
The current frontal image was always displayed beside the linked prior image when available, so that findings reported with temporal progression could be accurately identified.

%%%% Study-level quality control
% - Image quality issue
% - Wrong prior image
% - Report quality issue
% - Findings list issue
% - Unacceptable image/report
% - No issue
First, for each radiological study, annotators were presented with the original report in Spanish and the list of all extracted positive finding sentences in both Spanish and English (see \cref{fig:annotation_interface}A).
The study was assessed for any major issues with the image quality, report quality, selection of the prior image, or list of findings, as well as for the presence of unfiltered protected health information or paediatric patients.
Radiologists also discarded those studies for which the lateral projection was necessary to assess a finding.
% In those cases considered unsuitable for labelling, the radiologists indicated the type of issue for which they were rejected.
If any such issue was flagged, the study was removed from further annotation, and agreement between both radiologists was required to accept each case.
The 989 cases with no consensus were reevaluated by a group of three senior radiologists, requiring approval by at least two of them.
We additionally discarded 7 studies linked to rejected prior studies.

%%%% Box annotation: basic, priors

In the second stage, those images passing the quality control stage were annotated. %For this, each team member was assigned a random list of positive finding sentences from this dataset, marking with bounding boxes the individual finding indicated on the image. 
Each positive finding sentence was reviewed individually by two radiologists. If the finding was localisable, each annotator drew one or more tight bounding boxes covering the area(s) where it was visible in the image, as shown in \cref{fig:annotation_interface}B. In particular, for diffuse findings whose bounding box(es) would cover more than half of the image, annotators were instructed not to draw any boxes.

\subsection*{Annotation post-processing and arbitration}
%%%% Multi-rater adjudication
% \todo[inline]{@José Maria/Joaquín: We should explain how adjudication was done for tasks 1 and 2: number (and level) of annotators, review by senior(s), etc.}
% Multi-rater box arbitration criteria:
% \begin{enumerate}
%     \item (Exclude boxes with area $<0.1\%$)
%     \item Has boxes
%     \item Contained
%     \item Seniority
%     \item More boxes
%     \item Smaller boxes
%     \item Random
% \end{enumerate}

We filtered out individual boxes with areas smaller than 0.1\% of the respective image, likely corresponding to spurious clicks in the annotation interface.
Then, for each finding, we arbitrated which of the two annotations (i.e.~sets of boxes) should be treated as official. This was done by applying a sequence of binary criteria to both annotations until a unique choice could be made.
In order of priority, we preferred the annotation that:
(1)~has any boxes;
(2)~has at least 80\% of its area contained in the other, therefore assumed to be more precise;
(3)~was created by a senior radiologist;
(4)~has more boxes;
(5)~has smaller overall area.
All remaining 436 cases (7.0\%) were decided at random.
The unselected extra annotations are also released as part of the dataset, for researchers interested in studying inter-observer variability.
% (1)~prefer annotations with boxes;
% (2)~prefer annotations with at least 80\% of the area contained in the other, therefore assumed to be more precise;
% (3)~prefer annotations by senior radiologists over junior radiologists;
% (4)~prefer annotations with more boxes;
% (5)~prefer annotations with smaller overall area;
% (6)~select one annotation at random.

\subsection*{Assembly of grounded reports}

Each finding sentence was extracted along with a source span from the original report.
The positions of these excerpts in the text were then used to merge and sort the lists of positive and negative findings in the same order as originally reported.

In order to standardise the data for training machine learning models, we partitioned 70\% of studies for training, 10\% for validation, and the remaining 20\% for testing.
Splits were determined at random, stratified by the same finding categories as for data selection, and ensuring all studies of each patient were kept within a single partition.
This process was also applied to the remaining eligible studies outside of \datasetname, resulting in a stratified split for the entire PadChest dataset.

\section*{Results}

\subsection*{Dataset composition}

\begin{figure*}[tp]
    \centering
    \includegraphics[scale=0.55]{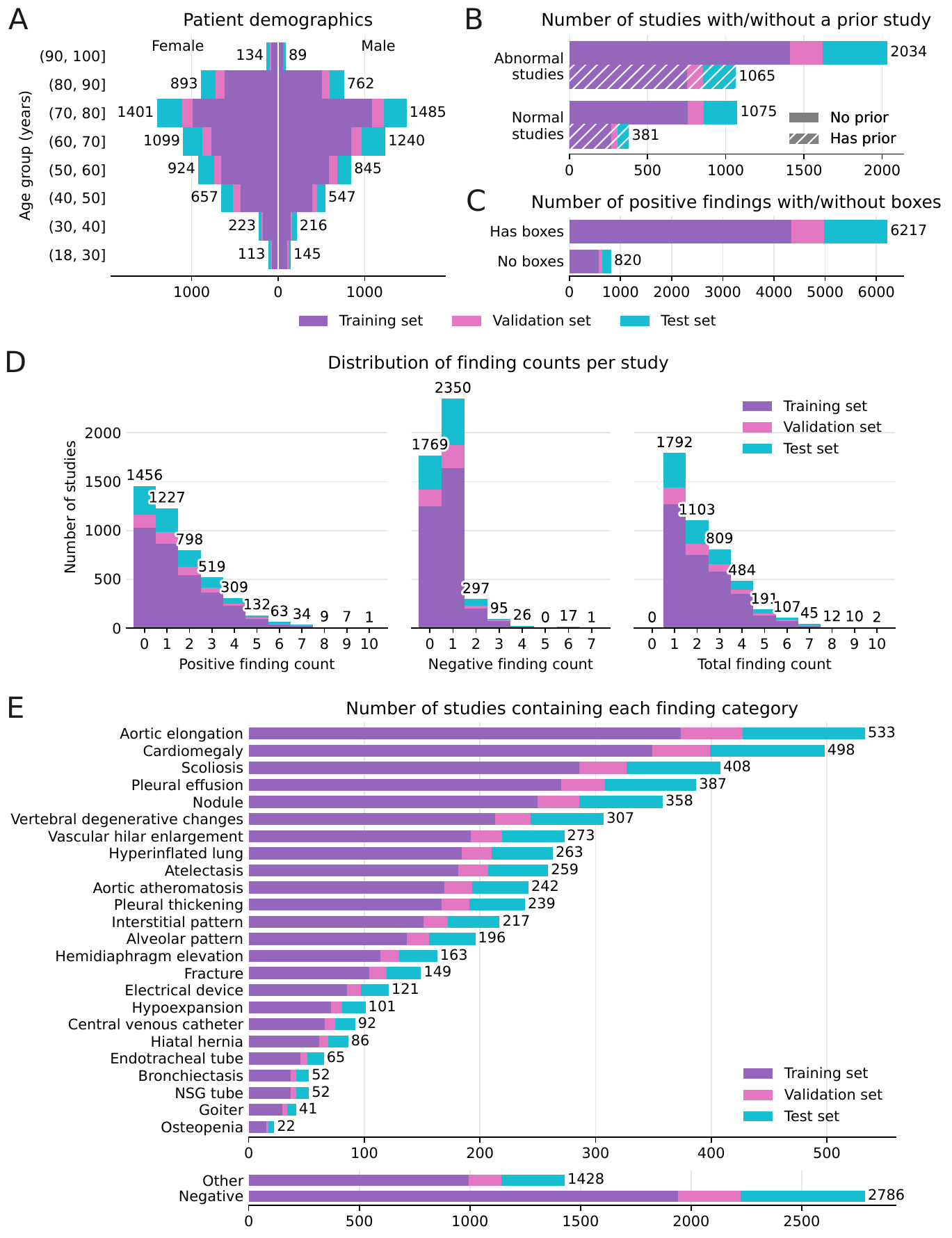}
    \caption{Descriptive statistics for the studies included in the \datasetname dataset.
    (Panel~A)~Joint distribution of patient sex and age.
    (Panel~B)~Availability of a prior study for abnormal and normal studies.
    (Panel~C)~Availability of manual box annotations per positive finding.
    (Panel~D)~Histograms of number of individual finding sentences (positive, negative, and total) per study.
    (Panel~E)~Distribution of the 24 finding categories, as well as studies including `Other' and negative findings.
    Note how all distributions of demographics, availability of prior studies, and finding counts are well represented across training, validation, and test partitions.
    }
    \label{fig:descriptive_stats}
\end{figure*}

The resulting \datasetname dataset is composed of 4,555 studies, of which 1,456 (32\%) are normal and 3,099 (68\%) abnormal. The median age is 69 years, and samples have an equal representation of men and women, with a balanced distribution between the two groups (see \cref{fig:descriptive_stats}A).
% All PadChest studies were originally acquired from consecutive patients between 2009 and 2017 at the University Hospital Sant Joan d'Alacant, Spain \cite{bustos2019padchest}.%\todo{Earlier years were automatically annotated. Selection of manually annotated cases explains the later 2014--2017 range}
% \wip{In particular, the eligible subset for \datasetname was all acquired in 2014--2017.}
The studies included in \datasetname were all acquired in 2014--2017, as this was the time range selected for manual annotation in PadChest \cite{bustos2019padchest}.
Prior images are available for 31.7\% of all studies. As expected, \cref{fig:descriptive_stats}B shows this proportion is higher among abnormal studies (34.4\%) than normal ones (26.2\%), since patients with positive findings are more likely to require follow-up scans for monitoring.

\datasetname includes 7,037 positive findings and 3,422 negative findings in total.
On average, the grounded report for each abnormal study contains 2.27 positive and 0.52 negative findings, whereas normal studies have 1.25 negative findings.
Detailed distributions are shown in \cref{fig:descriptive_stats}D.
We observe a long tail of positive finding counts, with 246 studies (5.4\%) containing five or more.
An overwhelming majority (84.4\%) of all 2,786 reports that include any negative findings has a single one, typically equivalent to ``no (other) significant findings''.
% Studies with a single negative finding:
% - for abnormal studies: 1095 / (3099 - 1769) = 82.3%
% - for normal studies: 1255 / 1456 = 86.2%

\Cref{fig:descriptive_stats}E shows the distribution of the main 24 finding categories, with a balanced representation across all partitions. %As described in Methods, the studies in \datasetname were both sampled and partitioned into training, validation, and test subsets with stratification according to these categories, to preserve the distribution from the original PadChest \cite{bustos2019padchest}, which is reflective of the routine clinical practice.
%Note how all distributions of demographics, availability of prior studies, and finding counts have balanced representation across the training, validation, and test partitions.

% Findings distribution is the same than in PadChest (link on Fig. 4) 

%\todo[inline]{Antonio: We need to comment these statistics, maybe someone with radiology background could add some thoughts about the distribution (if it makes sense or there is something that should be explained).}

Progression status (see \cref{fig:progression}) was available for 306 finding sentences (11.4\% of all positive findings) in studies with linked prior studies and 323 (7.4\%) in studies without. Combined, the evolution of these findings was as follows: 386 stable (61.3\%), 108 improving (17.2\%), 71 worsening (11.3\%), 56 new (8.9\%), and 8 resolved (1.3\%). The five most common finding categories with progression status were pleural effusion, nodule, alveolar pattern, interstitial pattern, and atelectasis. 

Location labels with anatomical regions were available for 64.3\% of the finding sentences. 
In addition to the selected list of finding categories used for stratification, radiologists also labelled a total of 1,956 finding sentences that were not included in the primary stratification criteria. These additional findings were labelled with bounding boxes because they were deemed clinically relevant or radiologically visible. The complete list of these additional entities, along with their respective counts of images and bounding boxes, is provided in \cref{tab:other_entities}, being the 4 most frequent entities chronic changes, %(n = 493), 
infiltrates, %(n = 154) 
fibrotic band % (n = 153) 
and increased density. %(n = 112).

\subsection*{Box annotations}
% \wip{\Cref{fig:boxes-per-finding} shows the average number of boxes annotated for each finding.}

% \begin{figure*}
%     \centering
%     \includegraphics[width=2.0\columnwidth]{figures/boxes-per-finding.pdf}
%     \caption{Average number of boxes labelled per finding category.}
%     \label{fig:boxes-per-finding}
% \end{figure*}

\begin{figure*}[tp]
    \centering
    \includegraphics[scale=0.5]{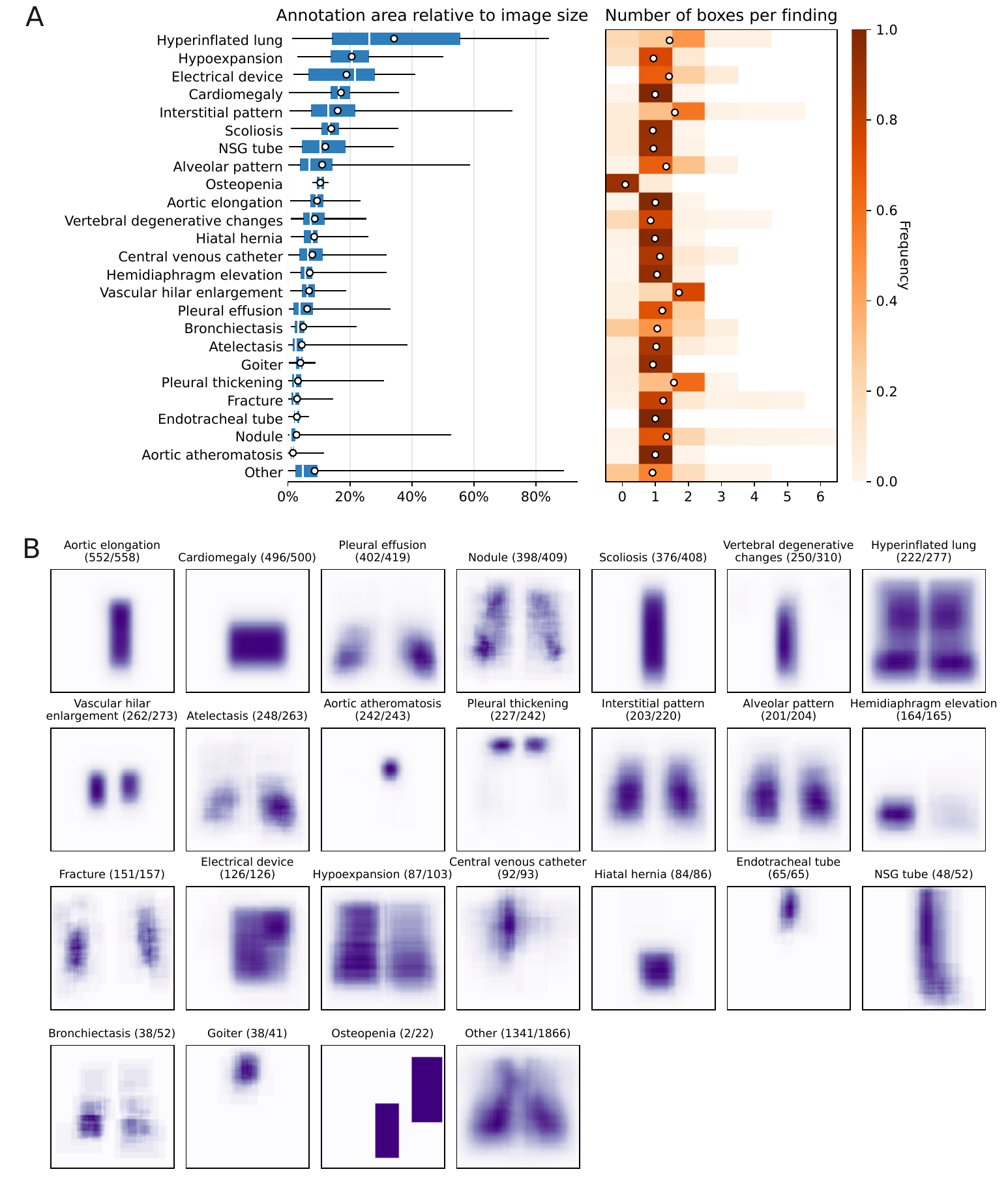}
    \caption{
    Descriptive statistics for the official box annotations included in the \datasetname dataset, stratified by finding category.
    (Panel~A)~Distributions of total area (box plots, left) and number of boxes (histograms, right) per annotation, sorted by mean area.
    Box plots indicate minimum/maximum, quartiles, median (vertical line), and mean (circle).
    Histograms additionally show the mean (circle).
    (Panel~B)~Heatmaps illustrating how often each pixel location was contained in a set of boxes for a finding in the given category.
    The numbers in parentheses indicate the counts of individual findings annotated with boxes and the total. Note that 224 studies contain two or more finding sentences in the same category.
    The scale of each heatmap is normalised independently for visualisation purposes.
    }
    \label{fig:box-stats}
\end{figure*}

% Comments:
% - Electrical device, NSG tube, CVC: can have large boxes covering the length of the tube or wires
Box annotation statistics are visualised in \cref{fig:box-stats}.
We observe that certain hardware categories---including electrical device (e.g.~pacemaker), nasogastric tube, and central venous catheter---tend to have relatively large boxes.
This is because the rectangular annotations cover the full horizontal and vertical extent of the long tubes and pacemaker wires, as evidenced by the spatial heatmaps in \cref{fig:box-stats}B.

% - Nodule, interstitial/alveolar pattern: variability between radiologists on drawing multiple tight boxes vs bigger encompassing boxes
It is also worth noting the variability in how radiologists annotated some finding categories, e.g.~nodules and interstitial/alveolar patterns.
At times, the latter were annotated with multiple small boxes precisely localising each instance in the image, and, in other cases, with larger boxes encompassing the whole affected areas. Up to 81.9\% of finding sentences have associated anatomical location labels (\cref{sec:locations}).

% - Empty annotations: occurs most often for diffuse/chronic findings like osteopenia, bronchiectasis, hyperexpansion, hypoinflation
% Lastly, note that not all positive findings were localised with boxes.
Lastly, note that 11.7\% of the total 7,037 positive findings were not localised with boxes, as indicated in \cref{fig:descriptive_stats}C.
Consistently with the annotation instructions, this happened particularly often for diffuse, chronic, or global finding categories, such as osteopenia, bronchiectasis, hyperexpansion, and hypoinflation.
Extra box annotations are available for 5,242 findings, with a median intersection-over-union of 0.530 compared to the official boxes.

% Stages: 
% Task 1: Report level QC. 3860 cases. Labels: image quality issue (214), wrong prior image (41), report quality issue (485), findings quality issue (269), no issue (3634), unacceptable image / report (57). 

% Task 2: Annotate the finding. 68% complete (8657 findings). For each finding, draw one or more boxes. 

%Images with no findings (389), 1 finding (3342), 2 findings (1611), 3 findings (331), 4 findings (147), 5 findings (26), 6 findings (10), 7 findings (2). 

\section*{Discussion}
%\todo[inline]{Kenji - Add 2-3 sentences on novelty and utility of using LLMs for data pre-processing}

% Limitations:
% \begin{itemize} -> INTEGRATED INTO THE TEXT
%    % \item Reports are not complete; they're written for the requested purpose
%    % \item Some findings like osteopenia are no longer reported from CXR alone (e.g. now require densitometry)
%    % \item Some diffuse or borderline findings are more susceptible to inter-rater variability
%     %\item Hyper/hypoinflation is clearer on lateral projection (shape of diaphragm)
%     %\item LLM processing is not perfect
% \end{itemize}

In this work, we present \datasetname, a first-in-class dataset designed to train and evaluate grounded report generation models from CXR images.
A key differentiator of \datasetname is the inclusion of comprehensive sentence-level bounding box annotations for all clinically relevant findings in each image, as judged by expert radiologists.
% This ensures the dataset provides a complete description of findings, aligning with our definition of a grounded report that describes all findings with accompanying localisation. Reports lacking comprehensiveness were filtered out during quality control, ensuring only CXRs with exhaustive annotations were included.
Samples with missing findings were filtered out during quality control, ensuring completeness of the grounded reports.

\datasetname offers several advantages that enhance the dataset's utility. First, it provides bilingual (Spanish and English) sentence-level findings alongside fine-grained standardised terminology, enhancing accessibility and utility across different linguistic contexts. Also, most bounding boxes are accompanied by location labels, providing multimodal spatial information for abnormalities. In addition, the dataset includes progression status and prior X-rays for a subset of studies, enabling research into disease evolution.
% By including detailed localization and comprehensive annotations of all clinically relevant findings, \datasetname provides a valuable resource for developing and evaluating \groundedreportgeneration models from CXR images.
Further, its comprehensive annotations with sentence-level structure support direct evaluation of logical, grounding, and spatial entailment using metrics such as RadFact \cite{bannur2024maira2}.

Inter-observer variability and subjectivity pose challenges for annotation, modelling, and evaluation---especially for borderline or diffuse findings. To address this, \datasetname includes annotations from two radiologists per finding, offering alternative perspectives and highlighting uncertainties.

%\todo[inline]{Reorder and consolidate limitations to clearly group into source data limitations (real-world context etc.) and annotation limitations.}
\subsection*{Limitations}

%Several limitations should also be acknowledged. 

We acknowledge some \datasetname limitations, most of them derived from the use of the original PadChest\cite{bustos2019padchest} data. First, although this dataset represents a large, retrospective cohort, it originates from a single hospital in Spain, likely containing biases related to regional healthcare practices, and does not fully represent diverse populations.

%First, although derived from a single hospital in Spain, the original PadChest dataset represents a large, retrospective cohort reflecting routine clinical practice, including a wide variety of pathologies. Detailed patient demographics are provided in the original publication \cite{bustos2019padchest}. However, the monocentric origin may introduce biases related to regional healthcare practices and population characteristics.

%Second, the annotations use rectangular bounding boxes, which may not precisely conform to the irregular shapes of anatomical structures and pathological findings. Consequently, they may include areas not part of the actual finding, affecting the precision of localization tasks. This limitation is common in medical imaging datasets but is notable for tasks requiring high spatial accuracy.

%Regarding quality, all images in \datasetname, acquired between 2014 and 2017, include many digitized from radiological films, resulting in lower image quality compared to modern standards. The images are in PNG format, with a reduced grayscale range compared to DICOM, limiting the visibility of subtle findings. Radiologists were also unable to adjust key parameters, such as window width and level, during annotation, affecting their ability to assess certain areas.

Regarding image quality, all studies were acquired between 2014 and 2017, with many digitized from radiological films rather than captured digitally, resulting in lower image quality compared to contemporary standards. The images are in PNG format with a reduced grayscale range compared to DICOM, %which offers higher bit-depth and contrast resolution. This can 
affecting the visibility of subtle findings, such as interstitial lung disease patterns or small pneumothoraces. Additionally, during annotation, radiologists were unable to adjust window width and level or apply image filters, affecting their assessment of certain areas, such as the pulmonary interstitium, retrocardiac regions, and the visualization of tubes and catheters.
Moreover, the cohort includes many bedridden and ICU patients, whose radiographs are of lower quality due to immobility and suboptimal positioning.  %introducing challenges for annotation. %While enhancing realism, this introduces challenges for annotation. % and model training, particularly regarding image quality.
% Nonetheless, although non-optimal image quality poses annotation challenges, models trained on such data are expected to be more robust in realistic scenarios.  %may generalize better and obtain higher accuracy in real scenarios.\todo{Highlight robustness to realistic data}
Nonetheless, non-optimal image quality reflects realistic conditions for training and testing robust models.

Regarding projection limitations, the dataset predominantly contains postero-anterior (PA) images without corresponding lateral views. Certain findings, such as vertebral fractures, hyperkyphosis, and pulmonary hyperinflation, are better assessed on lateral projections. The absence of lateral views meant that some findings in the reports could not be corroborated or annotated on the available images and were discarded, limiting the representation of pathologies that are projection-dependent.

Although LLMs have demonstrated impressive capabilities in de-identification \cite{xiao2023deidentification}, translation \cite{khanna2024multilingual}, and error detection in radiology reports \cite{gertz2024gpt4}, they are not without flaws. In our data processing, GPT-4 occasionally produced truncated sentences during automated extraction from reports. However, labels from the original PadChest dataset ensured that associated findings were annotated on the images. We also noted some literal translations such as `condensation' instead of `consolidation', though they are consistent across studies.

It is also worth noting that the time interval with prior radiographs available in PadChest was not always optimal for disease comparison, especially in cases of acute pathology. This inconsistency could limit the assessment of disease progression or the evaluation of temporal changes, which are important aspects in longitudinal studies.

%\todo[inline]{Consolidate limitations into narrative paragraph}
% To overcome those limitations, future efforts could focus on:

% - Enhancing Dataset Diversity: Incorporating data from multiple institutions and diverse populations to improve generalizability.
  
% - Improving Annotation Precision: Using more precise annotation methods, such as polygonal segmentation, to better conform to the shapes of anatomical structures and pathological findings.
  
% - Upgrading Image Quality: Including higher-resolution images in standard DICOM format to facilitate better visualization and analysis.
  
% - Expanding Projections: Collecting corresponding lateral views and other relevant imaging modalities to provide a more comprehensive assessment of findings.

\subsection*{Conclusions and future work}

%We conclude that \datasetname is a valuable resource specifically intended to advance CXR analysis by enabling the training and evaluation of models for automated grounded reporting. 
By providing comprehensive annotations of all clinically relevant findings along with their localisations, we hope that \datasetname will foster new avenues in medical imaging research and support the development of more robust and interpretable models in radiology.

To address the limitations mentioned, future efforts could focus on enhancing diversity by incorporating data from multiple institutions to improve generalisability. Additionally, %upgrading the image quality by 
including higher-resolution images in the standard DICOM format would facilitate more detailed visualization and analysis. Finally, expanding the dataset to include lateral views and other relevant imaging modalities would allow for a more comprehensive findings assessment.

%\todo[inline]{Add overall conclusions}

% Exclude back matter from word count: https://tex.stackexchange.com/a/259296
%TC:ignore
%\singlespacing
\footnotesize
\section*{Notes}

We thank the radiologists team from from Sant Joan d'Alacant Hospital---led by Joaquín Galant Herrero and composed by Maria Dolores Sánchez Valverde, Lara Jaques Pérez, Lourdes Pérez Rodríguez, Marife Lorente Fernández, Jorge Calbo Maiques, María Culiáñez Casas, Laila Santirso Abuelbar, Diego Mauricio Angulo Henao, Vicente Pedro Davó Quiñonero, María Begerano Fayos, Jorge Caballero Serra, Rocío Belda Márquez, and Carla Fuster Redondo---for their contributions with data annotation.
We further thank Fabian Falck, Fernando Pérez-García, and Harshita Sharma for their valuable feedback on the manuscript, Anton Schwaighofer for the software infrastructure support, Max Ilse for early contributions to the methodology, and Hannah Richardson for help navigating data compliance.

This work was led by the PadChest team and supported by project UA-MICROSOFT01 from the University of Alicante, funded by Microsoft Corporation and coordinated by Antonio Pertusa.  

\paragraph{CRediT author statement:}
D.C.C.: Conceptualization, Data curation, Formal analysis, Methodology, Project administration, Software, Visualization, Writing -- original draft, Writing -- review \& editing.
A.B.: Conceptualization, Formal analysis, Methodology, Software, Visualization, Writing -- original draft, Writing -- review \& editing.
S.B.: Data curation, Methodology, Software, Visualization, Writing -- review \& editing.
S.L.H.: Conceptualization, Data curation, Methodology, Software, Writing -- review \& editing.
K.B.: Data curation, Methodology, Software, Writing -- review \& editing.
M.T.W.: Data curation, Methodology, Writing -- review \& editing.
M.D.S.V., L.J.P., and L.P.R.: Investigation (data annotation).
K.T.: Conceptualization, Funding acquisition, Project administration, Writing -- original draft, Writing -- review \& editing.
J.M.S.: Conceptualization, Project administration, Resources.
J.A.V.: Conceptualization, Funding acquisition, Project administration, Resources, Writing -- review \& editing.
J.G.H.: Conceptualization, Investigation (data annotation), Methodology, Project administration, Writing -- review \& editing.
A.P.: Conceptualization, Funding acquisition, Project administration, Supervision,  Writing -- original draft, Writing -- review \& editing.

\normalsize

\singlespacing

\bibliographystyle{unsrt}
\bibliography{bibliography}

\begin{thebibliography}{10}

\bibitem{rajpurkar2023current}
Pranav Rajpurkar and Matthew~P Lungren.
\newblock The current and future state of ai interpretation of medical images.
\newblock {\em N Engl J Med}, 388(21):1981--1990, 2023.

\bibitem{yildirim2024multimodal}
Nur Yildirim, Hannah Richardson, Maria~Teodora Wetscherek, Junaid Bajwa, Joseph
  Jacob, Mark~Ames Pinnock, Stephen Harris, Daniel~C. Castro, Shruthi Bannur,
  Stephanie Hyland, Pratik Ghosh, Mercy Ranjit, Kenza Bouzid, Anton
  Schwaighofer, Fernando Pérez-García, Harshita Sharma, Ozan Oktay, Matthew
  Lungren, Javier Alvarez-Valle, Aditya Nori, and Anja Thieme.
\newblock Multimodal healthcare {AI}: Identifying and designing clinically
  relevant vision-language applications for radiology.
\newblock In {\em Proceedings of the 2024 CHI Conference on Human Factors in
  Computing Systems}, page 444, 2024.

\bibitem{bannur2024maira2}
Shruthi Bannur, Kenza Bouzid, Daniel~C. Castro, Anton Schwaighofer, Sam
  Bond-Taylor, Maximilian Ilse, Fernando Pérez-García, Valentina Salvatelli,
  Harshita Sharma, Felix Meissen, Mercy Ranjit, Shaury Srivastav, Julia Gong,
  Fabian Falck, Ozan Oktay, Anja Thieme, Matthew~P. Lungren, Maria~Teodora
  Wetscherek, Javier Alvarez-Valle, and Stephanie~L. Hyland.
\newblock {MAIRA-2}: Grounded radiology report generation, 2024.

\bibitem{bernstein2023incorrect}
Michael~H Bernstein, Michael~K Atalay, Elizabeth~H Dibble, Aaron~WP Maxwell,
  Adib~R Karam, Saurabh Agarwal, Robert~C Ward, Terrance~T Healey, and
  Grayson~L Baird.
\newblock Can incorrect artificial intelligence ({AI}) results impact
  radiologists, and if so, what can we do about it? {A} multi-reader pilot
  study of lung cancer detection with chest radiography.
\newblock {\em European Radiology}, 33(11):8263--8269, 2023.

\bibitem{moor2023foundation}
Michael Moor, Oishi Banerjee, Zahra Shakeri~Hossein Abad, Harlan~M. Krumholz,
  Jure Leskovec, Eric~J. Topol, and Pranav Rajpurkar.
\newblock Foundation models for generalist medical artificial intelligence.
\newblock {\em Nature}, 616(7956):259--265, April 2023.

\bibitem{bustos2019padchest}
Aurelia Bustos, Antonio Pertusa, Jose-Maria Salinas, and Maria De~La
  Iglesia-Vayá.
\newblock {PadChest}: A large chest x-ray image dataset with multi-label
  annotated reports.
\newblock {\em Medical Image Analysis}, 66:101797, 2020.

\bibitem{irvin2019chexpert}
Jeremy Irvin, Pranav Rajpurkar, Michael Ko, Yifan Yu, Silviana Ciurea-Ilcus,
  Chris Chute, Henrik Marklund, Behzad Haghgoo, Robyn~L. Ball, Katie
  Shpanskaya, Jayne Seekins, David~A. Mong, Safwan~S. Halabi, Jesse~K.
  Sandberg, Ricky Jones, David~B. Larson, Curtis~P. Langlotz, Bhavik~N. Patel,
  Matthew~P. Lungren, and Andrew~Y. Ng.
\newblock {CheXpert}: A large chest radiograph dataset with uncertainty labels
  and expert comparison.
\newblock In {\em Proceedings of the {AAAI} {Conference} on {Artificial}
  {Intelligence} ({AAAI} 2019)}, volume~33, pages 590--597. AAAI Press, July
  2019.

\bibitem{radgraphxl}
Jean-Benoit Delbrouck, Pierre Chambon, Zhihong Chen, Maya Varma, Andrew
  Johnston, Louis Blankemeier, Dave Van~Veen, Tan Bui, Steven Truong, and
  Curtis Langlotz.
\newblock {R}ad{G}raph-{XL}: A large-scale expert-annotated dataset for entity
  and relation extraction from radiology reports.
\newblock In {\em Findings of the Association for Computational Linguistics ACL
  2024}, pages 12902--12915, August 2024.

\bibitem{wang2017chestxray8}
Xiaosong Wang, Yifan Peng, Le~Lu, Zhiyong Lu, Mohammadhadi Bagheri, and
  Ronald~M. Summers.
\newblock {ChestX-ray8}: Hospital-scale chest {X}-ray database and benchmarks
  on weakly-supervised classification and localization of common thorax
  diseases.
\newblock In {\em Proceedings of the IEEE Conference on Computer Vision and
  Pattern Recognition (CVPR)}, pages 2097--2106. IEEE, 2017.

\bibitem{johnson2019mimiccxr}
Alistair E.~W. Johnson, Tom~J. Pollard, Seth~J. Berkowitz, Nathaniel~R.
  Greenbaum, Matthew~P. Lungren, Chih-ying Deng, Roger~G. Mark, and Steven
  Horng.
\newblock {MIMIC-CXR}, a de-identified publicly available database of chest
  radiographs with free-text reports.
\newblock {\em Scientific Data}, 6(1):317, December 2019.

\bibitem{demner2016iuxray}
Dina Demner-Fushman, Marc~D Kohli, Marc~B Rosenman, Sonya~E Shooshan, Laritza
  Rodriguez, Sameer Antani, George~R Thoma, and Clement~J McDonald.
\newblock Preparing a collection of radiology examinations for distribution and
  retrieval.
\newblock {\em Journal of the American Medical Informatics Association},
  23(2):304--310, 2016.

\bibitem{feng2021candidptx}
Sijing Feng, Damian Azzollini, Ji~Soo Kim, Cheng-Kai Jin, Simon~P. Gordon,
  Jason Yeoh, Eve Kim, Mina Han, Andrew Lee, Aakash Patel, Joy Wu, Martin
  Urschler, Amy Fong, Cameron Simmers, Gregory~P. Tarr, Stuart Barnard, and Ben
  Wilson.
\newblock Curation of the candid-ptx dataset with free-text reports.
\newblock {\em Radiology: Artificial Intelligence}, 3(6):e210136, 2021.

\bibitem{chambon2024chexpertplus}
Pierre Chambon, Jean-Benoit Delbrouck, Thomas Sounack, Shih-Cheng Huang,
  Zhihong Chen, Maya Varma, Steven~QH Truong, Chu~The Chuong, and Curtis~P.
  Langlotz.
\newblock {CheXpert Plus}: Augmenting a large chest {X}-ray dataset with text
  radiology reports, patient demographics and additional image formats, 2024.

\bibitem{Wu20}
Joy Wu, Yaniv Gur, Alexandros Karargyris, Ali~Bin Syed, Orest Boyko, Mehdi
  Moradi, and Tanveer Syeda-Mahmood.
\newblock Automatic bounding box annotation of chest x-ray data for
  localization of abnormalities.
\newblock In {\em 2020 IEEE 17th International Symposium on Biomedical Imaging
  (ISBI)}, pages 799--803, 2020.

\bibitem{Wu21}
Joy~T Wu, Nkechinyere Agu, Ismini Lourentzou, Ismini Lourentzou, Arjun Sharma,
  Joseph~Alexander Paguio, Jasper~Seth Yao, Edward~C Dee, William Mitchell,
  Satyananda Kashyap, Andrea Giovannini, Leo~Anthony Celi, and Mehdi Moradi.
\newblock {Chest ImaGenome} dataset for clinical reasoning, 2021.

\bibitem{Lanfredi22}
Ricardo~Bigolin Lanfredi, Joyce~D. Schroeder, and T.~Tasdizen.
\newblock Localization supervision of chest x-ray classifiers using
  label-specific eye-tracking annotation.
\newblock {\em Frontiers in radiology}, 3, 2022.

\bibitem{vindr}
Ha~Q Nguyen, Khanh Lam, Linh~T Le, Hieu~H Pham, Dat~Q Tran, Dung~B Nguyen,
  Dung~D Le, Chi~M Pham, Hang~TT Tong, Diep~H Dinh, et~al.
\newblock {VinDr-CXR}: An open dataset of chest {X}-rays with radiologist's
  annotations.
\newblock {\em Scientific Data}, 9(1):429, 2022.

\bibitem{boecking2022mscxr}
Benedikt Boecking, Naoto Usuyama, Shruthi Bannur, Daniel Coelho~de Castro,
  Anton Schwaighofer, Stephanie Hyland, Maria~Teodora Wetscherek, Tristan
  Naumann, Aditya Nori, Javier Alvarez~Valle, Hoifung Poon, and Ozan Oktay.
\newblock {MS-CXR}: Making the most of text semantics to improve biomedical
  vision-language processing (version 0.1), 2022.

\bibitem{bannur2023mscxrt}
Shruthi Bannur, Stephanie Hyland, Qianchu Liu, Fernando Pérez-García, Max
  Ilse, Daniel Coelho~de Castro, Benedikt Boecking, Harshita Sharma, Kenza
  Bouzid, Anton Schwaighofer, Maria~Teodora Wetscherek, Hannah Richardson,
  Tristan Naumann, Javier Alvarez~Valle, and Ozan Oktay.
\newblock {MS-CXR-T}: Learning to exploit temporal structure for biomedical
  vision-language processing (version 1.0.0), 2023.

\bibitem{ocae202}
Zhiyong Lu, Yifan Peng, Trevor Cohen, Marzyeh Ghassemi, Chunhua Weng, and Shubo
  Tian.
\newblock {Large language models in biomedicine and health: current research
  landscape and future directions}.
\newblock {\em Journal of the American Medical Informatics Association},
  31(9):1801--1811, 08 2024.

\bibitem{bodenreider2004umls}
Olivier Bodenreider.
\newblock The {U}nified {M}edical {L}anguage {S}ystem (umls): integrating
  biomedical terminology.
\newblock {\em Nucleic Acids Research}, 32(suppl1):D267--D270, January 2004.

\bibitem{mullenbach2018explainable}
James Mullenbach, Sarah Wiegreffe, Jon Duke, Jimeng Sun, and Jacob Eisenstein.
\newblock Explainable prediction of medical codes from clinical text.
\newblock In {\em Proceedings of the 2018 Conference of the North American
  Chapter of the Association for Computational Linguistics: Human Language
  Technologies, Volume 1 (Long Papers)}, volume~1, pages 1101--1111, 2018.

\bibitem{xiao2023deidentification}
Yuxin Xiao, Shulammite Lim, Tom~Joseph Pollard, and Marzyeh Ghassemi.
\newblock In the name of fairness: Assessing the bias in clinical record
  de-identification.
\newblock In {\em Proceedings of the 2023 ACM Conference on Fairness,
  Accountability, and Transparency}, pages 123--137, 2023.

\bibitem{khanna2024multilingual}
Praneet Khanna, Gagandeep Dhillon, Venkata Buddhavarapu, Ram Verma, Rahul
  Kashyap, and Harpreet Grewal.
\newblock Artificial intelligence in multilingual interpretation and radiology
  assessment for clinical language evaluation ({AI-MIRACLE}).
\newblock {\em Journal of Personalized Medicine}, 14(9):923, 2024.

\bibitem{gertz2024gpt4}
Roman~Johannes Gertz, Thomas Dratsch, Alexander~Christian Bunck, Simon
  Lennartz, Andra-Iza Iuga, Martin~Gunnar Hellmich, Thorsten Persigehl, Lenhard
  Pennig, Carsten~Herbert Gietzen, Philipp Fervers, David Maintz, Robert
  Hahnfeldt, Jonathan Kottlors, and Linda Moy.
\newblock Potential of {GPT-4} for detecting errors in radiology reports:
  Implications for reporting accuracy.
\newblock {\em Radiology}, 311(1):e232714, 2024.

\bibitem{openai2024gpt4}
OpenAI.
\newblock Gpt-4 technical report, 2024.

\end{thebibliography}

\appendix
\onecolumn
% Prefix reference labels with appendix section letter, e.g. Table B.1
\counterwithin{figure}{section}
\counterwithin{table}{section}
\counterwithin{lstlisting}{section}

\section{Finding label mapping for stratification}
\begingroup
\newlength{\groupcolwidth}
\setlength{\groupcolwidth}{3cm}
\footnotesize
\begin{longtable}{llrr}
\caption{Finding group labels used for stratification. Grouping of finding labels was done manually in conjunction with a consultant radiologist.}
\label{tab:group_labels} \\
\toprule
Group label & Finding label & Image count & Boxes count \\
\midrule
\endfirsthead
% \bottomrule
\endfoot
\caption{Group labels used for stratification (cont.)} \\
\toprule
Group label & Finding label & Image count & Boxes count \\
\midrule
\endhead
Negative & N/A & 1456 & 0 \\
\midrule
\multirow{5}{\groupcolwidth}{Aortic elongation} 
& aortic elongation & 464 & 465 \\
& aortic button enlargement & 38 & 33 \\
& descendent aortic elongation & 27 & 27 \\
& supra aortic elongation & 27 & 31 \\
& ascendent aortic elongation & 2 & 2 \\
\midrule
\multirow{2}{\groupcolwidth}{Cardiomegaly} 
& cardiomegaly & 498 & 495 \\
& pericardial effusion & 2 & 2 \\
\midrule
\multirow{12}{\groupcolwidth}{Nodule} 
& pseudonodule & 103 & 132 \\
& calcified granuloma & 100 & 139 \\
& nodule & 89 & 107 \\
& nipple shadow & 68 & 100 \\
& pulmonary mass & 24 & 24 \\
& granuloma & 20 & 28 \\
& multiple nodules & 13 & 33 \\
& end on vessel & 5 & 5 \\
& mass & 4 & 4 \\
& soft tissue mass & 4 & 4 \\
& miliary opacities & 2 & 4 \\
& pleural mass & 1 & 1 \\
\midrule
\multirow{7}{\groupcolwidth}{Pleural effusion} 
& pleural effusion & 208 & 264 \\
& costophrenic angle blunting & 191 & 226 \\
& minor fissure thickening & 16 & 14 \\
& loculated pleural effusion & 4 & 4 \\
& loculated fissural effusion & 3 & 3 \\
& fissure thickening & 2 & 2 \\
& major fissure thickening & 2 & 2 \\
\midrule
Scoliosis & scoliosis & 408 & 377 \\
\midrule
\multirow{3}{30mm}{Vertebral degenerative changes}
& vertebral degenerative changes & 269 & 263 \\
& vertebral anterior compression & 41 & 2 \\
& vertebral compression & 3 & 1 \\
\midrule
\multirow{3}{\groupcolwidth}{Hyperinflated lung}
& air trapping & 225 & 316 \\
& flattened diaphragm & 32 & 52 \\
& hyperinflated lung & 22 & 32 \\
\midrule
\multirow{3}{\groupcolwidth}{Vascular hilar enlargement}
& vascular hilar enlargement & 193 & 325 \\
& hilar congestion & 78 & 139 \\
& pulmonary artery enlargement & 2 & 4 \\
\midrule
\multirow{6}{\groupcolwidth}{Atelectasis} 
& laminar atelectasis & 154 & 164 \\
& atelectasis & 87 & 84 \\
& lobar atelectasis & 11 & 12 \\
& atelectasis basal & 6 & 5 \\
& segmental atelectasis & 4 & 4 \\
& total atelectasis & 1 & 1 \\
\midrule
Aortic atheromatosis & aortic atheromatosis & 243 & 243 \\
\midrule
\multirow{4}{\groupcolwidth}{Pleural thickening}
& apical pleural thickening & 205 & 333 \\
& pleural thickening & 20 & 25 \\
& calcified pleural thickening & 11 & 12 \\
& calcified pleural plaques & 6 & 8 \\
\midrule
% \pagebreak %%%%%%%%%%%%%%%% PAGE BREAK %%%%%%%%%%%%%%%%
\multirow{4}{\groupcolwidth}{Interstitial pattern}
& interstitial pattern & 195 & 311 \\
& ground glass pattern & 11 & 16 \\
& reticular interstitial pattern & 10 & 16 \\
& reticulonodular interstitial pattern & 4 & 7 \\
\midrule
\multirow{5}{\groupcolwidth}{Alveolar pattern}
& alveolar pattern & 146 & 211 \\
& consolidation & 42 & 43 \\
& cavitation & 14 & 15 \\
& air bronchogram & 2 & 2 \\
& abscess & 1 & 1 \\
\midrule
\multirow{5}{\groupcolwidth}{Electrical device}
& pacemaker & 104 & 147 \\
& dual chamber device & 35 & 53 \\
& single chamber device & 19 & 26 \\
& DAI (implantable cardioverter-defibrillator) & 14 & 20 \\
& electrical device & 4 & 4 \\
\midrule
\multirow{2}{\groupcolwidth}{Hemidiaphragm elevation}
& diaphragmatic eventration & 84 & 92 \\
& hemidiaphragm elevation & 81 & 81 \\
\midrule
\multirow{6}{\groupcolwidth}{Fracture} 
& callus rib fracture & 108 & 144 \\
& rib fracture & 22 & 25 \\
& humeral fracture & 13 & 13 \\
& clavicle fracture & 10 & 11 \\
& vertebral fracture & 3 & 0 \\
& fracture & 1 & 1 \\
\midrule
\multirow{2}{\groupcolwidth}{Hypoexpansion}
& volume loss & 75 & 68 \\
& hypoexpansion & 28 & 29 \\
\midrule
\multirow{4}{\groupcolwidth}{Central venous catheter}
& central venous catheter via jugular vein & 46 & 47 \\
& central venous catheter via subclavian vein & 17 & 24 \\
& central venous catheter & 15 & 19 \\
& reservoir central venous catheter & 15 & 16 \\
\midrule
\multirow{1}{\groupcolwidth}{Hiatal hernia}
& hiatal hernia & 86 & 85 \\
\midrule
\multirow{2}{\groupcolwidth}{Endotracheal tube}
& endotracheal tube & 39 & 39 \\
& tracheostomy tube & 26 & 26 \\
\midrule
\multirow{1}{\groupcolwidth}{NSG tube}
& NSG tube & 52 & 49 \\
\midrule
\multirow{1}{\groupcolwidth}{Bronchiectasis}
& bronchiectasis & 52 & 55 \\
\midrule
\multirow{1}{\groupcolwidth}{Goiter}
& goiter & 41 & 38 \\
\midrule
\multirow{2}{\groupcolwidth}{Osteopenia}
& osteopenia & 13 & 1 \\
& osteoporosis & 9 & 1 \\
\bottomrule
\end{longtable}
\normalsize
\endgroup

\clearpage
\section{Other finding types not used for stratification}
\footnotesize
\begin{longtable}{lrr}
\caption{Other finding types not used for stratification}
\label{tab:other_entities} \\
\toprule
Finding label & Image count & Boxes count \\
\midrule
\endfirsthead
\bottomrule
\endfoot
\caption{Other finding types not used for stratification (cont.)} \\
\toprule
Finding label & Image count & Boxes count \\
\midrule
\endhead

chronic changes & 493 & 223 \\
infiltrates & 154 & 171 \\
fibrotic band & 153 & 177 \\
increased density & 112 & 120 \\
kyphosis & 103 & 8 \\
sternotomy & 76 & 79 \\
suture material & 74 & 73 \\
hilar enlargement & 69 & 115 \\
metal & 55 & 60 \\
gynecomastia & 53 & 92 \\
calcified densities & 41 & 49 \\
mammary prosthesis & 41 & 73 \\
osteosynthesis material & 41 & 41 \\
bronchovascular markings & 33 & 40 \\
sclerotic bone lesion & 28 & 30 \\
tracheal shift & 27 & 27 \\
bullas & 25 & 35 \\
azygos lobe & 24 & 24 \\
mastectomy & 23 & 24 \\
superior mediastinal enlargement & 23 & 23 \\
mediastinic lipomatosis & 21 & 21 \\
mediastinal enlargement & 17 & 17 \\
vascular redistribution & 17 & 18 \\
axial hyperostosis & 15 & 10 \\
surgery breast & 15 & 17 \\
non axial articular degenerative changes & 14 & 19 \\
surgery & 14 & 12 \\
thoracic cage deformation & 13 & 15 \\
artificial heart valve & 10 & 9 \\
costochondral junction hypertrophy & 10 & 11 \\
pneumothorax & 10 & 8 \\
surgery neck & 10 & 10 \\
calcified adenopathy & 9 & 11 \\
adenopathy & 8 & 6 \\
mediastinal mass & 8 & 8 \\
surgery lung & 8 & 5 \\
chest drain tube & 7 & 7 \\
obesity & 7 & 1 \\
artificial mitral heart valve & 6 & 6 \\
central vascular redistribution & 6 & 7 \\
pectum excavatum & 6 & 0 \\
heart valve calcified & 5 & 5 \\
humeral prosthesis & 5 & 5 \\
air fluid level & 4 & 4 \\
cervical rib & 4 & 6 \\
kerley lines & 4 & 7 \\
pneumoperitoneo & 4 & 6 \\
abnormal foreign body & 3 & 3 \\
artificial aortic heart valve & 3 & 3 \\
catheter & 3 & 3 \\
lytic bone lesion & 3 & 3 \\
prosthesis & 3 & 3 \\
sternoclavicular junction hypertrophy & 3 & 3 \\
subacromial space narrowing & 3 & 4 \\
subcutaneous emphysema & 3 & 3 \\
surgery heart & 3 & 3 \\
aortic aneurysm & 2 & 2 \\
aortic endoprosthesis & 2 & 2 \\
azygoesophageal recess shift & 2 & 2 \\
blastic bone lesion & 2 & 4 \\
calcified fibroadenoma & 2 & 2 \\
cyst & 2 & 2 \\
hydropneumothorax & 2 & 2 \\
lung vascular paucity & 2 & 2 \\
surgery humeral & 2 & 2 \\
Chilaiditi sign & 1 & 1 \\
endoprosthesis & 1 & 1 \\
gastrostomy tube & 1 & 1 \\
mediastinal shift & 1 & 1 \\
pectum carinatum & 1 & 0 \\
ventriculoperitoneal drain tube & 1 & 1 \\
\end{longtable}
\normalsize

\clearpage
\section{Language model prompting details}
\label{app:llm_details}

We employed OpenAI GPT-4 \cite{openai2024gpt4} v0613 via a private instance of the Microsoft Azure OpenAI service.
The processing was done in a structured way to extract single-finding sentences, translate them to English, link them to the existing PadChest labels, and copy each source sentence to allow reconstructing full grounded reports in the right order.
The expected output formats were specified by including the respective JSON schema in the model prompt, and outputs were validated to ensure consistency of the extracted metadata, using the Pydantic Python library.
GPT-4 worked more reliably than GPT-3.5~Turbo for these highly structured tasks and it was capable of processing all sentences simultaneously and without in-context examples (for positive findings extraction).
For extracting negative findings, 11 in-context examples were provided.

\begin{lstlisting}[
    caption={GPT-4 prompt for extracting positive finding phrases.},
    label={lst:positive_findings_prompt}
]
# System
You are an AI radiology assistant who is fluent in English and Spanish. You are helping process radiology reports from chest X-rays.

Note that, in Spanish, lung lobes are referred by abbreviations, for example: "LSD" for "lóbulo superior derecho" (right upper lobe).

You will receive a report in Spanish prefixed by "REPORT:". For different sentences in the report, you will receive a list of finding types prefixed by "SENTENCE FINDINGS:", and possibly a list of location labels prefixed by "SENTENCE LOCATIONS:", if available.

Instructions:
- You must process the sentence labels in the same order as given in the query.
- For each group of labels, extract verbatim the sentence from the report that describes them.
- For each finding label, provide a standalone phrase describing specifically the requested finding. It should be as close as possible to the full original sentence and not mention any other findings. Include as much context from the report as necessary for the phrase to make sense on its own.
- All finding labels must be processed, even if they are redundant.
- For each extracted sentence and generated phrase, also include an accurate translation in English.
- If finding labels are provided, identify which of these correspond to each finding. All of the provided location labels must be exactly matched to at least one finding in the same sentence, even if they are redundant.
- If a finding is reported with a comparison to a prior exam, classify the progression of the condition as "new", "worsening", "stable", "improving", or "resolved". If there is no comparison, set progression to "N/A".

The output should be formatted as a JSON instance that conforms to the JSON schema below.

As an example, for the schema {"properties": {"foo": {"title": "Foo", "description": "a list of strings", "type": "array", "items": {"type": "string"}}}, "required": ["foo"]}
the object {"foo": ["bar", "baz"]} is a well-formatted instance of the schema. The object {"properties": {"foo": ["bar", "baz"]}} is not well-formatted.

Here is the output schema:
```
{json_schema}
```

# User
REPORT: {report_es}
----
SENTENCE FINDINGS: {sentence_labels[0].finding}
SENTENCE LOCATIONS: {sentence_labels[0].locations}
----
SENTENCE FINDINGS: {sentence_labels[1].findings}
SENTENCE LOCATIONS: {sentence_labels[1].locations}
----
...
\end{lstlisting}

\begin{lstlisting}[
    caption={GPT-4 instruction for extracting negative finding phrases.},
    label={lst:negative_findings_prompt}
]
You are an AI radiology assistant. You are helping process reports from chest X-rays.

Please extract phrases from the radiology report which refer to objects, findings, or anatomies visible in a chest X-ray, or the absence of such.

Rules:
- If a sentence describes multiple findings, split them up into separate sentences.
- Exclude clinical speculation or interpretation (e.g. "... highly suggestive of pneumonia").
- Exclude recommendations (e.g. "Recommend a CT").
- Exclude comments on the technical quality of the X-ray (e.g. "there are low lung volumes").
- Include mentions of change (e.g. "Pleural effusion has increased") because change is visible when we compare two X-rays.
- If consecutive sentences are closely linked such that one sentence can't be understood without the other one, process them together.
- Mark whether the phrase refers to something normal ("non_finding"), abnormal ("finding"), or ambiguous.

The objective is to extract phrases which refer to things which can be located on a chest X-ray, or confirmed not to be present.
If the report is in a language other than English, provide an English translation of each sentence and phrase in the fields "orig_en" and "new_en".
\end{lstlisting}

\begin{lstlisting}[
    caption={One of the in-context examples provided to GPT-4 for extracting negative finding phrases. Given the `\texttt{findings\_text}', GPT-4 outputs the `\texttt{parsed\_report}'.},
    label={lst:negative_findings_fewshot}
]
"findings_text": "RX tórax posteroanterior:  Impresión diagnóstica: Sin hallazgos patológicos para la edad del paciente.",
"parsed_report":
[
    {
        "orig": "RX tórax posteroanterior:",
        "orig_en": "Chest X-ray posteroanterior:",
        "new": [""],
        "new_en": [""]
    },
    {
        "orig": "Impresión diagnóstica:",
        "orig_en": "Diagnostic impression:",
        "new": [""],
        "new_en": [""]
    },
    {
        "orig": "Sin hallazgos patológicos para la edad del paciente.",
        "orig_en": "No pathological findings for the patient's age.",
        "new": ["Sin hallazgos patológicos para la edad del paciente.|non_finding"],
        "new_en": ["No pathological findings for the patient's age.|non_finding"]
    }
]
\end{lstlisting}

\clearpage
\section{Progression status}
% \begin{figure*}[h]
%     \centering
%     \includegraphics[width=1\linewidth]{figures/top_10_labels_with_progression_status_stacked.png}
%     \caption{Distribution of progression statuses across the 10 top findings with prior studies available. This visualisation highlights the proportion of each progression category within specific medical findings.}
%     \label{fig:progression}
%     \todo[inline]{Daniel: Re-generate in consistent style using Aurelia's script}
% \end{figure*}
\begin{figure*}[h]
    \centering
    \includegraphics[width=\textwidth]{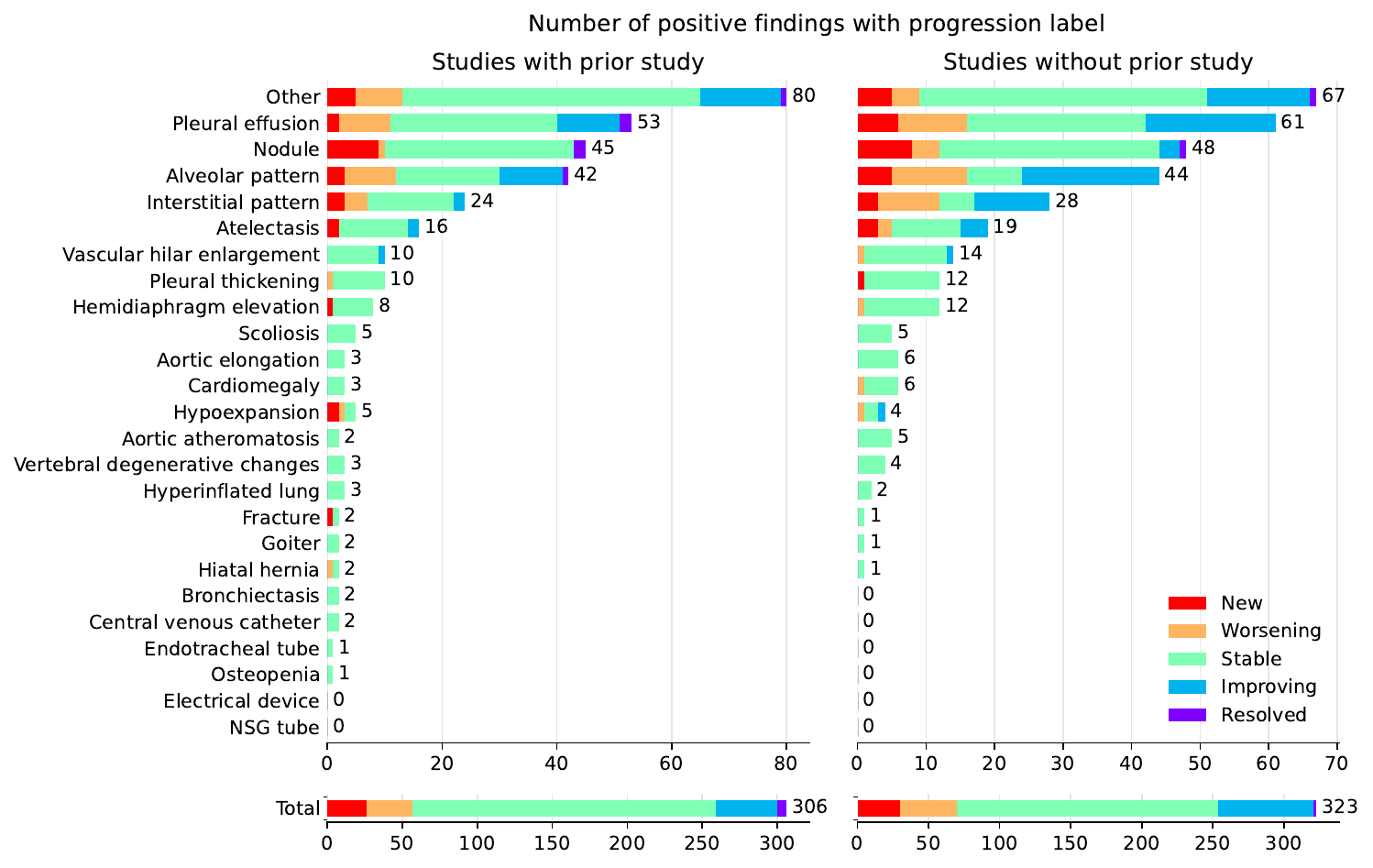}
    \caption{Distribution of progression statuses across finding categories, for studies with and without prior studies available.}
    \label{fig:progression}
\end{figure*}

\clearpage
\section{Location labels}
\label{sec:locations}

\captionof{table}{Distribution of location labels.}
\begin{tabular}{ll}
\toprule
\begin{minipage}[t]{0.45\textwidth}
\renewcommand{\DTstyle}{\normalfont}
\setlength{\DTbaselineskip}{1,96ex}
\footnotesize
\dirtree{%
.1 2691 images, 6588 boxes.
.2 extracorporal: 0 images, 0 boxes.
.2 right: 845 images, 1207 boxes.
.2 left: 645 images, 896 boxes.
.2 bilateral: 419 images, 929 boxes.
.2 cervical: 34 images, 39 boxes.
.2 soft tissue: 231 images, 420 boxes.
.3 subcutaneous: 8 images, 8 boxes.
.3 axilar: 31 images, 39 boxes.
.3 pectoral: 176 images, 358 boxes.
.4 nipple: 67 images, 136 boxes.
.2 bone: 445 images, 541 boxes.
.3 shoulder: 90 images, 114 boxes.
.4 acromioclavicular joint: 5 images, 8 boxes.
.4 rotator cuff: 6 images, 7 boxes.
.4 supraspisnous: 3 images, 4 boxes.
.4 humerus: 71 images, 82 boxes.
.5 humeral head: 28 images, 32 boxes.
.5 humeral neck: 4 images, 4 boxes.
.5 glenohumeral joint: 7 images, 8 boxes.
.3 clavicle: 18 images, 18 boxes.
.3 scapula: 10 images, 15 boxes.
.3 costoesternal: 2 images, 3 boxes.
.3 column: 146 images, 111 boxes.
.4 intersomatic space: 6 images, 6 boxes.
.4 dorsal vertebrae: 85 images, 54 boxes.
.4 cervical vertebrae: 4 images, 4 boxes.
.4 paravertebral: 20 images, 20 boxes.
.3 rib: 165 images, 238 boxes.
.4 anterior rib: 13 images, 15 boxes.
.4 posterior rib: 29 images, 39 boxes.
.4 rib cartilage: 4 images, 8 boxes.
.2 hemithorax: 145 images, 232 boxes.
.2 extrapleural: 3 images, 3 boxes.
.2 extrapulmonary: 1 images, 1 boxes.
.2 pleural: 471 images, 840 boxes.
.2 subpleural: 3 images, 7 boxes.
.2 fissure: 27 images, 27 boxes.
.3 minor fissure: 17 images, 17 boxes.
.3 major fissure: 3 images, 4 boxes.
.2 lobar: 358 images, 505 boxes.
.3 upper lobe: 191 images, 284 boxes.
.4 left upper lobe: 82 images, 116 boxes.
.5 lingula: 30 images, 34 boxes.
.4 right upper lobe: 93 images, 125 boxes.
.3 lower lobe: 101 images, 135 boxes.
.4 left lower lobe: 45 images, 58 boxes.
.4 right lower lobe: 43 images, 52 boxes.
.3 middle lobe: 44 images, 52 boxes.
.2 subsegmental: 41 images, 47 boxes.
.2 bronchi: 91 images, 112 boxes.
.2 peribronchi: 1 images, 2 boxes.
.3 diffuse bilateral: 31 images, 61 boxes.
.3 basal bilateral: 111 images, 194 boxes.
.2 paratracheal: 23 images, 33 boxes.
.2 airways: 191 images, 223 boxes.
.3 tracheal: 104 images, 113 boxes.
.3 bronchi: 91 images, 112 boxes.
}
\end{minipage}
&
\begin{minipage}[t]{0.55\textwidth}
\renewcommand{\DTstyle}{\normalfont}
\setlength{\DTbaselineskip}{1,96ex}
\footnotesize
\dirtree{%
.1 (continued).
.2 lung field: 1485 images, 2698 boxes.
.3 upper lung field: 297 images, 502 boxes.
.4 upper lobe: 275 images, 461 boxes.
.5 right upper lobe: 93 images, 125 boxes.
.5 apical: 161 images, 289 boxes.
.5 suprahilar: 10 images, 11 boxes.
.3 middle lung field: 526 images, 934 boxes.
.4 aortopulmonary window: 4 images, 3 boxes.
.4 hilar: 448 images, 800 boxes.
.5 pulmonary artery: 2 images, 1 boxes.
.5 hilar bilateral: 19 images, 31 boxes.
.5 perihilar: 73 images, 148 boxes.
.4 minor fissure: 17 images, 17 boxes.
.3 lower lung field: 916 images, 1325 boxes.
.4 basal: 330 images, 460 boxes.
.4 lower lobe: 101 images, 135 boxes.
.5 left lower lobe: 45 images, 58 boxes.
.5 right lower lobe: 43 images, 52 boxes.
.4 middle lobe: 44 images, 52 boxes.
.4 infrahilar: 54 images, 72 boxes.
.4 lingula: 30 images, 34 boxes.
.4 supradiaphragm: 2 images, 4 boxes.
.4 diaphragm: 211 images, 261 boxes.
.4 infradiaphragm: 34 images, 31 boxes.
.4 cardiophrenic angle: 20 images, 24 boxes.
.4 costophrenic angle: 196 images, 244 boxes.
.5 right costophrenic angle: 56 images, 57 boxes.
.5 left costophrenic angle: 85 images, 89 boxes.
.5 bilateral costophrenic angle: 4 images, 10 boxes.
.2 central: 86 images, 101 boxes.
.2 mediastinum: 1120 images, 1594 boxes.
.3 superior mediastinum: 143 images, 177 boxes.
.4 carotid artery: 0 images, 0 boxes.
.4 brachiocephalic veins: 1 images, 1 boxes.
.4 supra aortic: 24 images, 30 boxes.
.4 aortic button: 37 images, 38 boxes.
.4 superior cave vein: 54 images, 62 boxes.
.4 subclavian vein: 32 images, 44 boxes.
.3 lower mediastinum: 559 images, 625 boxes.
.4 anterior mediastinum: 2 images, 0 boxes.
.5 thymus: 0 images, 0 boxes.
.4 middle mediastinum: 494 images, 541 boxes.
.5 cardiac: 493 images, 539 boxes.
.6 coronary: 3 images, 3 boxes.
.4 posterior mediastinum: 67 images, 78 boxes.
.5 retrocardiac: 66 images, 77 boxes.
.3 aortic: 620 images, 743 boxes.
.2 esophageal: 3 images, 5 boxes.
.2 paramediastinum: 4 images, 5 boxes.
.2 paracardiac: 10 images, 15 boxes.
.2 epigastric: 1 images, 1 boxes.
.2 gastric chamber: 3 images, 3 boxes.
.2 hypochondrium: 16 images, 23 boxes.
.3 right hypochondrium: 13 images, 19 boxes.
.4 gallbladder: 5 images, 7 boxes.
.3 left hypochondrium: 3 images, 4 boxes.
}
\end{minipage}\\
\bottomrule
\end{tabular}
%\input{supplement/location_labels}
%TC:endignore

\end{document}